\def\year{2019}\relax
\documentclass[letterpaper]{article} 
\usepackage{aaai19}  
\usepackage{times}  
\usepackage{helvet}  
\usepackage{courier}  
\usepackage{url}  
\usepackage{graphicx}  
\usepackage{times}
\usepackage{latexsym}
\usepackage{graphicx}
\usepackage{url}
\usepackage{CJK}
\usepackage{xcolor}
\usepackage{multirow}
\usepackage{subfigure}
\usepackage{amssymb}
\usepackage{enumitem}
\usepackage{amsmath}
\newcommand{\citet}[1]{\citeauthor{#1}~\shortcite{#1}}

\begin{document}
 \begin{CJK*}{UTF8}{gbsn}
%
\title{Subword Encoding in Lattice LSTM for Chinese Word Segmentation}
\author{ 
Jie Yang, Yue Zhang, Shuailong Liang\\
Singapore University of Technology and Design\\
jieynlp@gmail.com, yue.zhang@wias.org.cn, shuailong\_liang@mymail.sutd.edu.sg
}
\maketitle
\begin{abstract}
We investigate a lattice LSTM network for Chinese word segmentation (CWS) to utilize words or subwords. It integrates the character sequence features with all subsequences information matched from a lexicon. The matched subsequences serve as information shortcut tunnels which link their start and end characters directly. Gated units are used to control the contribution of multiple input links. Through formula derivation and comparison, we show that the lattice LSTM is an extension of the standard LSTM with the ability to take multiple inputs. Previous lattice LSTM model takes word embeddings as the lexicon input, we prove that subword encoding can give the comparable performance and has the benefit of not relying on any external segmentor. The contribution of lattice LSTM comes from both lexicon and pretrained embeddings information, we find that the lexicon information contributes more than the pretrained embeddings information through controlled experiments. Our experiments show that the lattice structure with subword encoding gives competitive or better results with previous state-of-the-art methods on four segmentation benchmarks. Detailed analyses are conducted to compare the performance of word encoding and subword encoding in lattice LSTM. We also investigate the performance of lattice LSTM structure under different circumstances and when this model works or fails.
\end{abstract}

\section{Introduction}
Different from the English-like languages whose words are separated naturally, it is necessary to segment character sequence as word sequence in many East Asian languages, such as Chinese. Chinese word segmentation (CWS) has been thoughtfully studied from statistical methods to recent deep learning approaches. Most of them formalize CWS as a sequence labeling problem \cite{xue2003chinese}.

Neural network based Chinese word segmentation has attracted significant research attention due to its ability of non-linear feature representation and combination. Typical neural CWS models utilize the Long Short-Term Memory (LSTM) \cite{hochreiter1997long} or Convolution Neural Network (CNN) \cite{lecun1989backpropagation} as feature extractor, and take the character unigram and bigram embeddings as inputs \cite{pei2014max,yang2017neural}. Those models already achieve state-of-the-art performance on many CWS benchmarks \cite{zhou2017word,wang2017convolutional,yang2017neural}.

It has been shown that word information is beneficial to word segmentation \cite{zhang2007chinese,zhang2016transition,cai2016neural}. \citet{zhang2007chinese} built a transition-based word segmentor by utilizing hand-crafted word features, \citet{zhang2016transition} and \citet{cai2016neural} extended the transition framework as neural models which utilize the word embeddings. 

\begin{figure}[!tp] 
  \centering 
  \includegraphics[width=3.3in]{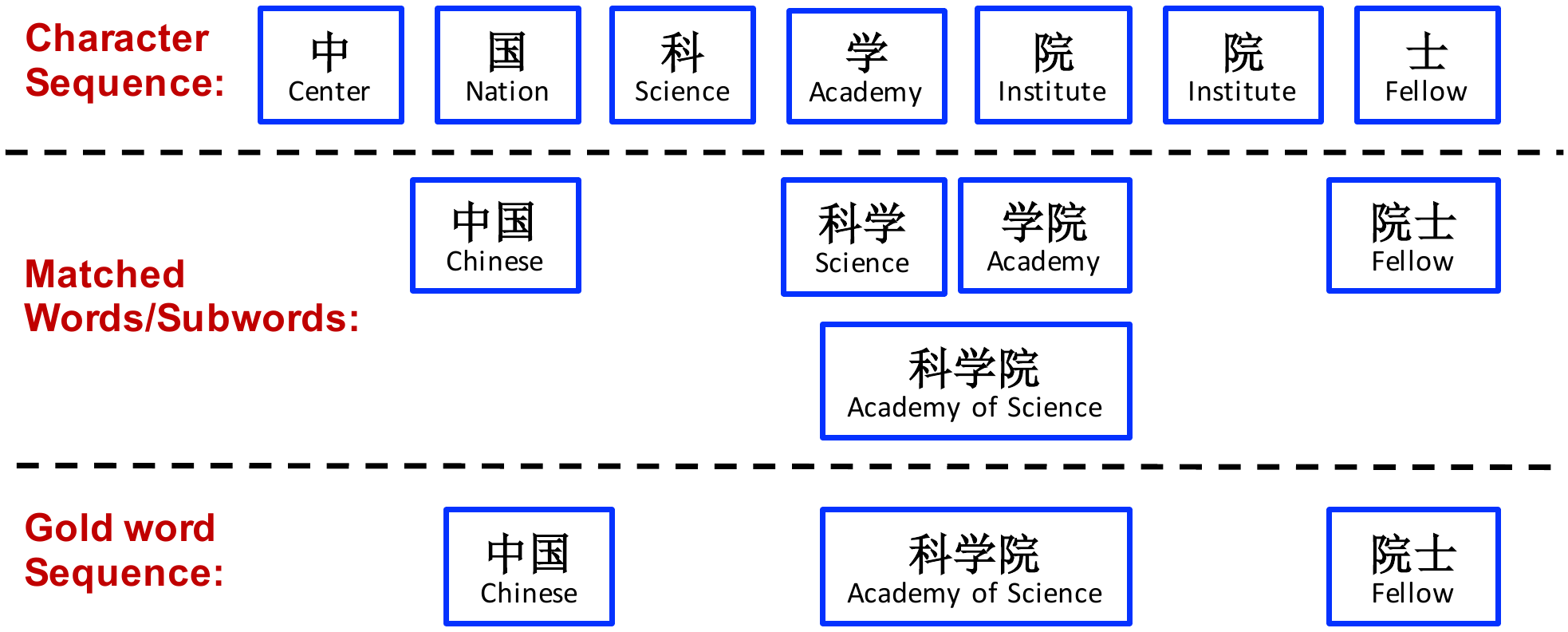}
  \caption{Segmentation with ambiguous words.}
  \label{fig:demofigure}
\end{figure}

One limitation of the above word-based models, however, is that word information can be utilized only for readily recognized words, namely those that are already in the output candidates. However, ambiguous words in a context can provide additional information for disambiguation. For instance, in Figure \ref{fig:demofigure}, the word ``科学院(Academy of Sciences)'' and ``学院(Academy)'' can be useful for determining the correct segmentation, which is ``科学院/(Academy of Sciences/)'', despite that ``学院(Academy)'' is not in the correct output. 

\begin{figure*}[!tp] 
  \centering 
  \includegraphics[width=7in]{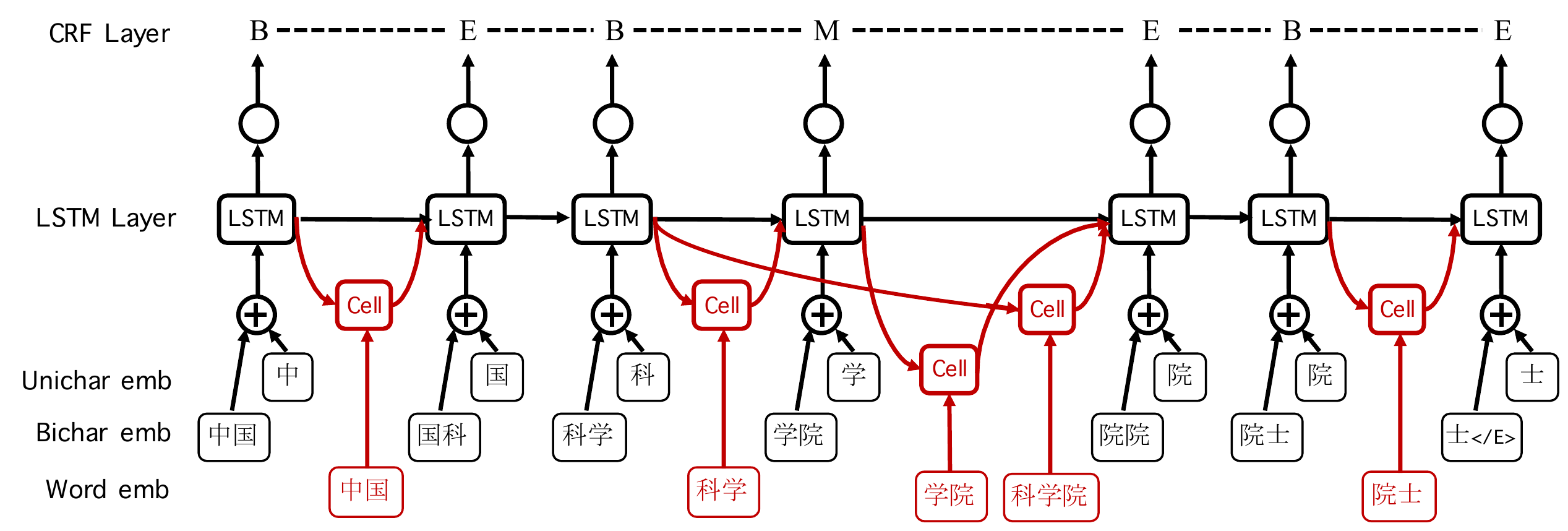}
  \caption{Models. Only forward LSTM is illustrated here.}
  \label{fig:structure}
\end{figure*}
\citet{zhang18chinese} proposed a lattice LSTM structure which can utilize the ambiguous words information in the named entity recognition (NER) task. The lattice structure is based on character LSTM sequence but leverages word information by using extra ``shortcut paths'' to link the memory cell between the start and the end characters of the word. To control the contribution of each ``shortcut path'', gated recurrent unit is used in each path. The final memory cell of character is the weighted sum of all the ``shortcut paths''. The ``shortcut paths'' are constructed by directly matching the sentence with a word lexicon, while the word lexicon comes from auto-segmented text. In this way, the lattice LSTM based NER system requires the segmentor information, although in an indirect way.

In this work, we extend the lattice LSTM structure in \citet{zhang18chinese} by using subword encoding which does not rely on any external segmentor\footnote{Our code is released at \url{https://github.com/jiesutd/SubwordEncoding-CWS}.}. We further examine the lattice LSTM which utilizes ambiguous words or subwords information on CWS tasks. Different from \citet{zhang18chinese} which takes the word embeddings as the lattice input, we additionally examine the Byte Pair Encoding (BPE) \cite{gage1994new} algorithm to encode the subword and construct the lattice LSTM with subword embeddings. The construction of subword embeddings does not rely on large segmented text which is necessary when building word embeddings. 

To our knowledge, we are the first to use the BPE for word segmentation. The comparison between word and subword embeddings are thoroughly studied. The contributions of word/subword lexicon and their pretrained embeddings are also investigated through controlled experiments. Experiments on four benchmarks show that the subword encoding in lattice LSTM can give comparable results with word embeddings, and they both can achieve state-of-the-art segmentation performance. In the end, we analyze two segmentation examples which show failed case for word encoding for lattice LSTM (``Lattice+Word'') and subword encoding for lattice LSTM (``Lattice+Subword''), respectively.

\section{Related Work}
Statistical word segmentation has been studied for decades \cite{sproat1996stochastic}. State-of-the-art models are either using the sequence labeling methods e.g. CRF \cite{lafferty2001conditional} with character features \cite{peng2004chinese,zhao2006effective} or taking the transition-based models with word features \cite{zhang2007chinese,sun2010word}. 

Similarly, neural word segmentors replace the hand-crafted features with neural representations. \citet{chen2015gated} and \citet{chen2015long} build neural CRF segmentors with GRU and LSTM to extract the representation on character embeddings, respectively. \citet{cai2016neural} and \citet{zhang2016transition} directly score the word sequences with beam search by utilizing both word information and character embeddings, both show comparable results with neural CRFs. \citet{yang2017neural} extend the word-based neural segmentor through pretraining the character representations with multi-task training on four external tasks, and observe significant improvement. \citet{zhou2017word} improve the segmentor with better character embeddings which include pre-segmented word context information through a new embedding training method on a large auto-segmented corpus.

Lattice RNNs have been used to model speech tokenization lattice \cite{sperber2017neural} and multi-granularity segmentation for NMT \cite{su2017lattice}. \citet{zhang18chinese} proposed a lattice LSTM for Chinese NER. It integrates the character sequence features and all matched word embeddings into a sequence labeling model, leading to a powerful NER system. \citet{N16-1106} proposed a DAG-structured LSTM structure which is similar to the lattice LSTM model, the DAG-LSTM binarizes the paths in the merging process but lattice LSTM merges the paths using gate controlled summation. \citet{chen2017dag} also built a DAG-LSTM structure for word segmentation. Different from their model which uses no memory cell in the word path, our lattice LSTM assigns one memory cell for each word path and merges them in the end character of the word. In addition, our model consistently gives better performance.

BPE is a data compression algorithm which iteratively merges the most frequent pair of bytes in a sequence as a new byte. In this work, we use BPE algorithm to merge characters rather than bytes in the text corpus, constructing the subwords which represent the most frequent character compositions in corpus level. It has been successfully used in neural machine translation by capturing the most frequent subwords instead of words \cite{sennrich2016neural}.

\section{Models}
We take the state-of-the-art LSTM-CRF framework as our baseline. For an input sentence with $m$ characters $s=c_1,c_2,\ldots,c_m$, where $c_i$ denotes the $i$th character, the segmentor is to assign each character $c_i$ with a label $l_i$, where $l_i\in \{B,M,E,S\}$ \cite{xue2003chinese}. The label $B, M$, $E$ and $S$ represent the \textit{begin, middle, end} of a word and single character word, respectively. Figure \ref{fig:structure} shows the segmentor framework on input character sequence ``中国科学院院士 \;(Fellow of the Chinese Academy of Sciences)'', where the black part represents the baseline LSTM-CRF model and the red part shows the lattice structure.

\subsection{Embedding Layer}
As shown in Figure \ref{fig:structure}, for each input character $c_i$, the corresponding character unigram embeddings and character bigram embeddings are represented as $\textbf{e}_{c_i}$ and $\textbf{e}_{c_ic_{i+1}}$, respectively. The character representation is calculated as following:

\begin{equation}
\textbf{x}_i = \textbf{e}_{c_i} \oplus \textbf{e}_{c_ic_{i+1}},
\end{equation}
where $\oplus$ represents \textit{concatenate} operation.

Unlike \citet{zhang2016transition} which uses a window to strengthen the local features, or \citet{zhou2017word} which adds a non-linear layer before the LSTM layer, we feed the character representation $(\textbf{x}_1,\textbf{x}_2,\ldots,\textbf{x}_m)$ into a bidirectional LSTM directly.

\subsection{Baseline LSTM Layer}
LSTM \cite{hochreiter1997long} is an advanced recurrent neural network (RNN) with extra memory cells which are used to keep the long-term information and alleviate the gradient vanishing problem. Equation \ref{eq:clstm} shows the calculation of $\overrightarrow{\textbf{h}}_i$ which is the forward LSTM representation of character $c_i$ .

\begin{equation}\label{eq:clstm}
\begin{aligned}
{\left[ \begin{array}{c}
\textbf{o}_i\\
\textbf{f}_i\\
\boldsymbol{\widetilde{c}}_i
\end{array} 
\right ]} 
&=
{\left[ \begin{array}{c}
\sigma\\
\sigma\\
tanh 
\end{array} 
\right ]} 
\Big(
\textbf{W}^{\top}
{\left[ \begin{array}{c}
\textbf{x}_i \\
\overrightarrow{\textbf{h}}_{i-1}\\ 
\end{array} 
\right ]} 
 +
\textbf{b}
 \Big)
\\
\textbf{i}_i &= \textbf{1} - \textbf{f}_i \\
\textbf{c}_i &= \textbf{f}_i\odot \textbf{c}_{i-1} + \textbf{i}_i \odot \boldsymbol{\widetilde{c}}_i \\
\overrightarrow{\textbf{h}}_i &= \textbf{o}_i\odot tanh(\textbf{c}_i) \\
\end{aligned}
\end{equation}
where $\textbf{i}_i, \textbf{f}_i$ and $\textbf{o}_i$ denote a set of input, forget and output gates, respectively. We choose the coupled LSTM structure \cite{greff2017lstm} which sets the input gate $\textbf{i}_i = \textbf{1}-\textbf{f}_i$. $\textbf{c}_i$ is the memory cell of character $c_i$. $\textbf{W}^{\top}$ and $\textbf{b}$ are model parameters. $\sigma()$ represents the sigmoid function.

For each input sentence $(\textbf{x}_1,\textbf{x}_2,\ldots,\textbf{x}_m)$, we calculate both the forward and backward LSTM representation as follows: 

\begin{equation}
\begin{array}{c}
\overrightarrow{\textbf{h}}_1, \overrightarrow{\textbf{h}}_2, \ldots, \overrightarrow{\textbf{h}}_m = \overrightarrow{LSTM}(\textbf{x}_1,\textbf{x}_2,\ldots,\textbf{x}_m) \\
\overleftarrow{\textbf{h}}_1, \overleftarrow{\textbf{h}}_2, \ldots, \overleftarrow{\textbf{h}}_m = \overleftarrow{LSTM}(\textbf{x}_1,\textbf{x}_2,\ldots,\textbf{x}_m),
\end{array} 
\end{equation}
where $\overrightarrow{LSTM}$ and $\overleftarrow{LSTM}$ represent the forward and backward LSTM, respectively. To incoperate the information from both sides, the hidden vector of character $c_i$ is the concatenation of the representations in both directions:

\begin{equation}
\textbf{h}_i = \overrightarrow{\textbf{h}}_i \oplus\overleftarrow{\textbf{h}}_i
\end{equation} 

A CRF layer (Eq. \ref{eq:crf}) is used on top of the hidden vectors $(\textbf{h}_1,\textbf{h}_2,\ldots,\textbf{h}_m)$ to perform label prediction.

\subsection{Lattice LSTM Layer} \label{sec:lattice}
The lattice LSTM adds ``shortcut paths'' (red part in Figure \ref{fig:structure}) to LSTM. The input of the lattice LSTM model is character sequence and all subsequences which are matched words in a lexicon $\mathbb{D}$. $\mathbb{D}$ is collected from auto-segmented Gigaword corpus or BPE encoding. Following \citet{zhang18chinese}, we use $w_{b,e}$ to represent the subsequence that has a start character index $b$ and a end character index $e$, and the embeddings of the subsequence is represented as $\textbf{e}_{w_{b,e}}$.

During the forward lattice LSTM calculation, the ``cell'' in Figure \ref{fig:structure} of a subsequence $w_{b,e}$ takes the hidden vector of the start character $\overrightarrow{\textbf{h}}_b$ and the subsequence (word or subword) embeddings $\textbf{e}_{w_{b,e}}$ as input, an extra LSTMcell (without output gate) is applied to calculate the memory vector of the sequence $\textbf{c}_{w_{b,e}}$:

\begin{equation}
\begin{aligned}
{\left[ \begin{array}{c}
\textbf{i}_{b,e} \\
\textbf{f}_{b,e}\\
\boldsymbol{\widetilde{c}}_{b,e}
\end{array} 
\right ]} 
&=
{\left[ \begin{array}{c}
\sigma \\
\sigma\\
tanh 
\end{array} 
\right ]} 
\Big(
\textbf{W}_s^{\top}
{\left[ \begin{array}{c}
\textbf{e}_{w_{b,e}} \\
\overrightarrow{\textbf{h}}_b\\ 
\end{array} 
\right ]} 
 +
\textbf{b}_s
 \Big)
\label{eq:word}\\
\textbf{c}_{b,e} &= \textbf{f}_{b, e}\odot \textbf{c}^c_b + \textbf{i}_{b, e} \odot \boldsymbol{\widetilde{c}}_{b, e} \\
\end{aligned}
\end{equation}
where $\textbf{c}_{b,e}$ is the memory cell of the shortcut path starting from character $c_b$ to character $c_e$. $\textbf{W}_s^{\top}$ and $\textbf{b}_s$ are model parameters of the shortcut path LSTM. Different from the standard LSTMcell which calculates both memory and output vectors, we calculate only the memory cell of the shortcut path. 

The subsequence output memory vector $\textbf{c}_{b,i}$ links to the end character $c_i$ as the input to calculate the hidden vector $\overrightarrow{\textbf{h}}_i$ of $c_i$. For character $c_i$ with multiple subsequence memory cell inputs\footnote{e.g. The first ``院(College)'' in Figure \ref{fig:structure} takes two subsequence memory vectors of both  ``学院(Academy)'' and ``科学院(Academy of Sciences)'' as input.}, we define the input set as $\mathbb{C}_i = \{\textbf{c}_{b,i}|b \in\{b'|w_{b',i} \in \mathbb{D}\}\}$ , we assign a unique gate for each subsequence input to control its contribution:

\begin{equation}
 \label{eq:gw}\\
\textbf{i}_{b,i} = \sigma
\big(
\textbf{W}^{g\top}{\left[ \begin{array}{c}
\textbf{x}_{i}\\
\textbf{c}_{b,i}\\ 
\end{array} 
\right ]} 
+\textbf{b}^g
\big) \\
\end{equation}
where $\textbf{W}^{g\top}$ and $\textbf{b}^g$ are model parameters for the gate. 

Until now, we have calculated the subsequence memory inputs $\mathbb{C}_i$ and their control gates $\mathbb{I}_i = \{\textbf{i}_{b,i}|b \in\{b'|w_{b',i} \in \mathbb{D}\}\}$. Following the idea of coupled LSTM \cite{greff2017lstm} which keeps the sum of input and forget gate as $\textbf{1}$, we normorize all the subsequence gates $\mathbb{I}_i$ with the standard LSTM input gate $\textbf{i}_i$ to ensure their sum equals to $\textbf{1}$ (Eq. \ref{eq:norm}).
\begin{equation} \label{eq:norm}
\begin{aligned}
\boldsymbol\alpha_{b,i} &= \dfrac{exp(\textbf{i}_{b,i})}{exp(\textbf{i}_i)+\sum\limits_{\textbf{i}_{b',i}\in \mathbb{I}_i}exp(\textbf{i}_{b',i})} \\
\boldsymbol\alpha_i &= \dfrac{exp(\textbf{i}_i)}{exp(\textbf{i}_i)+\sum\limits_{\textbf{i}_{b',i}\in \mathbb{I}_i}exp(\textbf{i}_{b',i})} \\
\end{aligned}
\end{equation}

$\boldsymbol\alpha_{b,i}$ and $\boldsymbol\alpha_i$ are the subsequence memory gate and the standard LSTM input gate after the normalization, respectively. The final forward lattice LSTM representation $\overrightarrow{\textbf{h}}_i$ of character $c_i$ is calculated as:

\begin{equation}\label{eq:latticelstm}
\begin{aligned}
{\left[ \begin{array}{c}
\textbf{o}_i\\
\textbf{f}_i\\
\boldsymbol{\widetilde{c}}_i
\end{array} 
\right ]} 
&=
{\left[ \begin{array}{c}
\sigma \\
\sigma\\
tanh 
\end{array} 
\right ]} 
\Big(
\textbf{W}^{\top}
{\left[ \begin{array}{c}
\textbf{x}_i \\
\overrightarrow{\textbf{h}}_{i-1}\\ 
\end{array} 
\right ]} 
 +
\textbf{b}
 \Big)
\\
\textbf{i}_i &= \textbf{1} - \textbf{f}_i \\
\textbf{c}_i &=  \sum\limits_{\boldsymbol{c}_{b,i} \in \mathbb{C}_i} \boldsymbol\alpha_{b,i}\odot \boldsymbol{c}_{b,i} + \boldsymbol\alpha_i\odot \boldsymbol{\widetilde{c}}_i  \\
\overrightarrow{\textbf{h}}_i &= \textbf{o}_i\odot tanh(\textbf{c}_i) \\
\end{aligned}
\end{equation}
where $\textbf{W}^{\top}$ and $\textbf{b}$ are the model parameters which are the same with the standard LSTM in Eq. \ref{eq:clstm}. Compare with Eq. \ref{eq:clstm}, Eq. \ref{eq:latticelstm} has a more complex memory calculation step which integrates both the standard character LSTM memory $\boldsymbol{\widetilde{c}}_i$ and all the matched subsequence memory inputs $\mathbb{C}_i$. In this respect, we can regard the lattice LSTM as an extension of the standard LSTM with the ability of taking multiple inputs.

The backward lattice LSTM representation $\overleftarrow{\textbf{h}}_i$ has a symmetrical calculation process as above. To give a fair comparison with the baseline bi-directional LSTM structure, we use the bi-directional lattice LSTM whose final hidden vector $\textbf{h}_i$ is the concatenation of the hidden vectors on both lattice LSTM directions. The same CRF layer as the baseline is used on top of the lattice LSTM layer.

\begin{table}[!tp]
\begin{center}
\begin{tabular}{|l|l|l|l|l|}
\hline 
\textbf{Dataset} &\textbf{Type}&\textbf{Train} & \textbf{Dev} & \textbf{Test} \\ 
\hline
\multirow{3}*{CTB6} 
&Sentence& 23.4k  &2.08k   &2.80k  \\
&Word& 641k  &59.9k   &81.6k  \\
&Char &1.06m &100k  &134k \\
\hline
\multirow{3}*{PKU} 
&Sentence&17.2k   & 1.91k &1.95k  \\
&Word&1.01m   &99.9k   &104k  \\
&Char &1.66m  &164k &173k \\
\hline
\multirow{3}*{MSR} 
&Sentence&78.2k   &8.69k   &3.99k  \\
&Word&2.12m   & 247k  &107k  \\
&Char &3.63m  & 417k  &184k  \\
\hline
\multirow{3}*{Weibo} 
&Sentence&20.1k   &2.05k   & 8.59k \\
&Word& 421k  &43.7k   & 188k \\
&Char &689k    &73.2k  &316k   \\
\hline
\end{tabular}
\end{center}
\caption{Statistics of datasets.}
\label{tab:OverallSta}
\end{table}

\subsection{Decoding and Training}
A standard CRF layer is used. The probability of a label sequence $y=l_1,l_2,\ldots,l_m$ is

\begin{equation}\label{eq:crf}
P(y|s)=\dfrac{exp(\sum\limits_{i=1}^m(F(l_i) + L(l_{i-1},l_i)))}{\sum\limits_{y^\prime\in \mathbb{C}(s)}exp(\sum\limits_{i=1}^m(F(l^\prime_i) + L(l^\prime_{i-1},l^\prime_i))},
\end{equation}
where $\mathbb{C}(s)$ is the set of all possible label sequences on sentence $s$ and $y^\prime$ is an arbitary label sequence. $F(l_i)=\textbf{W}^{l_i}\textbf{h}_i + b^{l_i}$  is the emission score from hidden vector $\textbf{h}_i$ to label $l_i$. $L(l_{i-1},l_i)$ is the transition score from $l_{i-1}$ to $l_i$. $\textbf{W}^{l_i}$ and $b^{l_i}$ are model parameters specific to label $l_i$.

Viterbi algorithm \cite{viterbi1967error} is used to decode the highest scored label sequence over the input sequence. During training, we choose sentence-level log-likelihood as the loss function.
\begin{equation}
 Loss = \sum\limits_{i=1}^{N}log(P(y_i|s_i)),
\end{equation}
where $y_i$ is the gold labels of sentence $s_i$.

\section{Experiments}
\subsection{Experimental Settings}
\textbf{Data}. We take the Chinese Treebank 6.0 (CTB6) \cite{xue2005penn} as our main dataset and split the train/dev/test following \citet{zhang2016transition}. We also evaluate our model on another three standard Chinese word segmentation datasets: PKU, MSR, and Weibo. PKU and MSR are taken from the SIGHAN 2005 bake-off \cite{emerson2005second} with standard data split. Different from the CTB6/PKU/MSR which are mainly based on formal news text, Weibo dataset is collected from informal social media in the NLPCC 2016 shared task \cite{qiu2016overview}, the standard split is used. Table \ref{tab:OverallSta} shows the details of the four investigated datasets.

\noindent \textbf{Hyperparameters}. Table \ref{tab:hyperparameter} shows the chosen hyperparameters in our model. We do not tune the hyperparameters based on each dataset but keep them same among all datasets. Standard gradient descent (SGD) with a learning rate decay is used as the optimizer. The embedding sizes of character unigram/bigram and word/subword are all in 50 dimensions. Dropout \cite{srivastava2014dropout} is used on both the character input and the word/subword input to prevent overfitting.

\begin{table}[!tp]
\begin{center}
\resizebox{\columnwidth}{!}{%
\begin{tabular}{|l|l||l|l|}
\hline \textbf{Parameter} &  \textbf{Value} & \textbf{Parameter} & \textbf{Value}\\ \hline
char emb size & 50 & bigram emb size & 50 \\
word emb size &50 &subword emb size &50\\
char dropout & 0.5 & lattice dropout & 0.5\\
LSTM layer &1  &LSTM hidden & 200\\
learning rate \textit{lr} & 0.01 &\textit{lr} decay & 0.05\\
\hline
\end{tabular}
}
\end{center}
\caption{Hyper-parameter values.}
\label{tab:hyperparameter}
\end{table}

\noindent \textbf{Embeddings}. We take the same character unigram embeddings, bigram embeddings and word embeddings with \citet{zhang2016transition}, which pretrain those embeddings using word2vec \cite{mikolov2013efficient} on Chinese Gigaword corpus\footnote{\url{https://catalog.ldc.upenn.edu/LDC2011T13}.}. The vocabulary of subword is constructed with 200000 merge operations and the subword embeddings are also trained using word2vec \cite{heinzerling2018bpemb}. Trie structure \cite{fredkin1960trie} is used to accelerate the building of lattice (matching words/subwords). All the embeddings are fine-tuned during training.

\begin{figure}[!tp] 
  \centering 
  \includegraphics[width=3.3in]{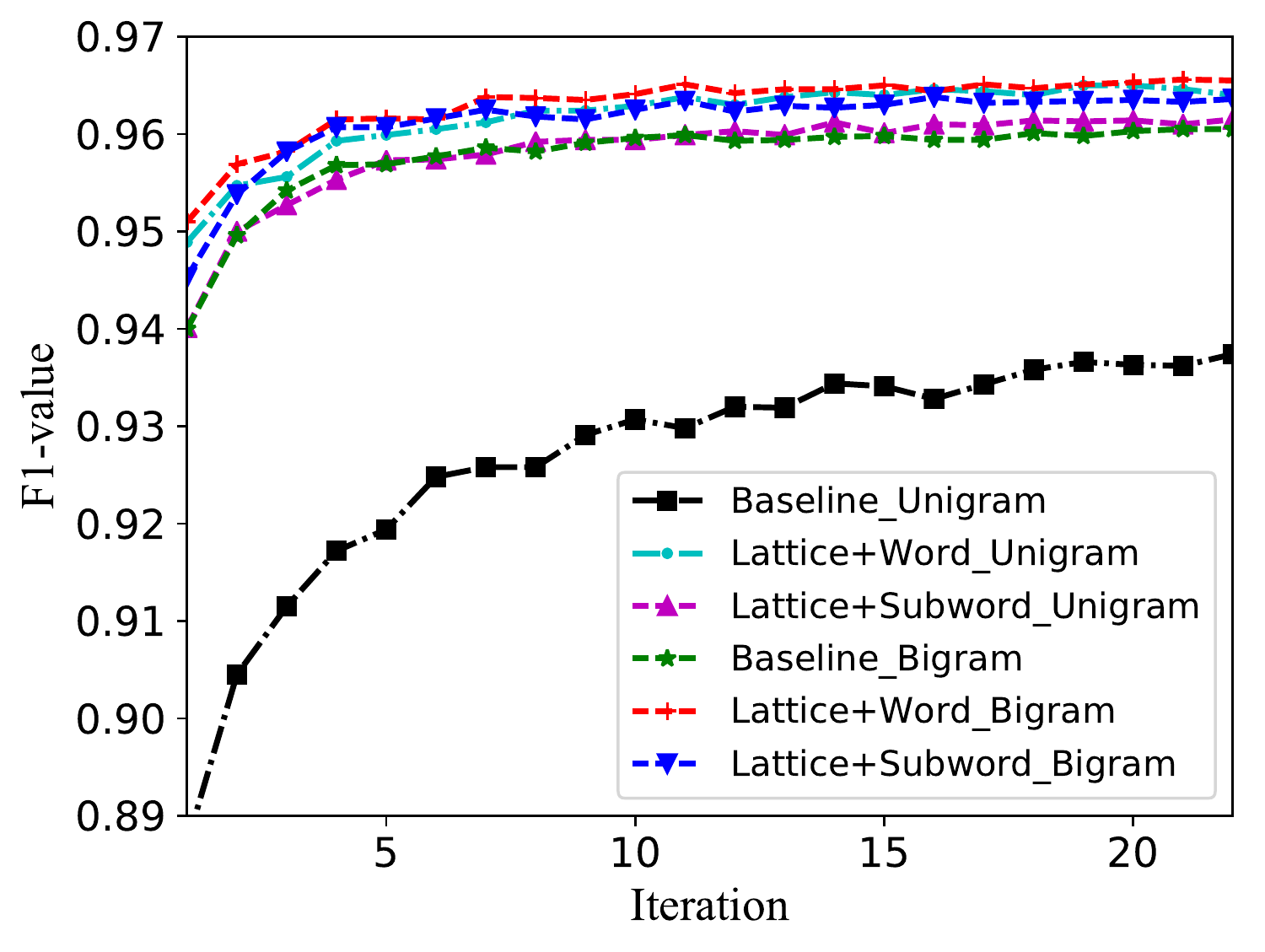}
  \caption{F1-value against training iterations.}
  \label{fig:iter}
\end{figure}

\subsection{Development Experiments}
We perform development experiments on CTB6 development dataset to investigate the contribution of character bigram information and the word/subword information. Figure \ref{fig:iter} shows the iteration curve of F-scores against different numbers of training iterations with different character representations. ``\_Unigram'' means model using only character unigram information and ``\_Bigram'' represents the model using both character unigram and bigram information (concatenating their embeddings). The performance of the baseline with only character unigram information is largely behind the others. By integrating the character bigram information, baseline model has a significant improvement. If we replace the character bigram with ``Lattice+Word'' or ``Lattice+Subword'', the model performance is even better. This proves that our lattice LSTM structure has a better ability to disambiguate the characters than character bigram information, we attribute this ability to the gate control mechanism which filters the noisy words/subwords. 

\citet{zhang18chinese} observed that character bigram information has a negative effect on ``Lattice+Word'' structure on Chinese NER task, while it is different on Chinese segmentation task. We find the character bigram information gives significant improvements on both ``Lattice+Word'' and ``Lattice+Subword'' structures. This is likely because character bigrams are informative but ambiguous, they can provide more useful character disambiguation evidence in segmentation task than in NER task where ``Lattice+Word'' already works well in disambiguating characters.

\begin{table}[!tp]
\begin{center}
\resizebox{\columnwidth}{!}{%
\begin{tabular}{|l|l|l|l|l|}
\hline 
\multirow{2}*{\textbf{Models}} &  \multirow{2}*{\textbf{CTB6}} & \multicolumn{2}{|c|}{\textbf{SIGHAN}} &\multirow{2}*{\textbf{Weibo}}\\
\cline{3-4}
   &   & \textbf{PKU} & \textbf{MSR} &\\ 
\hline
\citet{zheng2013deep} &-- &92.4 & 93.3 &--\\
\citet{pei2014max} &-- &95.2 &97.2  &--\\
\citet{ma2015accurate}&-- &95.1 &96.6  &--\\
\citet{zhang2016transition}* &96.0 &95.7 &97.7  &--\\
\citet{xu2016dependency} &95.8 & 96.1 &96.3  &--\\ 
\citet{cai2017fast} &-- & 95.8 &97.1  &--\\ 
\citet{yang2017neural}* &\textbf{96.2} &\textbf{96.3}  &97.5    &\textbf{95.5} \\
\citet{zhou2017word} &\textbf{96.2} & 96.0 & \textbf{97.8}  & --\\
\hline
\hline
Baseline &95.8 &95.3 &97.4&95.0\\
Lattice+Word &\textbf{96.3} &\textbf{95.9} &97.7 &95.1 \\
Lattice+Subword &96.1 &95.8 &\textbf{97.8}  &\textbf{95.3}\\
\hline
\end{tabular}
}
\end{center}
\caption{Main results (F1). * represents model utilizing external supervised information.  }
\label{tab:mainresults}
\end{table}

\subsection{Results}
We evaluate our model on four datasets. The main results and the recent state-of-the-art neural CWS models are listed in Table \ref{tab:mainresults}. \citet{zhang2016transition} integrated both discrete features and neural features in a transition-based framework. \citet{xu2016dependency} proposed the dependency-based gated recursive neural network to utilize long distance dependencies.  \citet{yang2017neural} utilized the character representations which are jointly trained on several tasks such as punctuation prediction, POS tagging, and heterogeneous segmentation task. \citet{zhou2017word} trained the character embeddings by including the segmentation label information from large auto-segmented text. While our ``Lattice+Subword'' does not need those steps. These works are orthogonal to and can be integrated to our lattice LSTM model. 

As shown in Table \ref{tab:mainresults}, the lattice LSTM models have significant improvement from the baseline on all datasets. The ``Lattice+Word'' model gives 11.9\%, 12.8\%, 11.5\%, 2.0\% error reductions on CTB6/PKU/MSR/Weibo datasets, respectively. And ``Lattice+Subword'' model has 7.14\%, 10.6\%, 15.4\%, 6.0\% error reductions on the four datasets, respectively. The reason for the different error reductions on different datasets is discussed in the analysis section. ``Lattice+Word'' and ``Lattice+Subword'' gives the best performance than other models on CTB6 and MSR, respectively. In PKU dataset, our lattice LSTM is slightly behind the model of \citet{yang2017neural,zhou2017word,xu2016dependency}, while the first two models utilize the external supervised or semi-supervised information and \citet{xu2016dependency} preprocess the dataset by replacing all the Chinese idioms, leading the comparison not entirely fair. In conclusion, both word encoding and subword encoding can help the lattice LSTM model gives comparable performance to the state-of-the-art word segmentation models.

``Lattice+Subword'' works better than ``Lattice+Word'' on MSR and Weibo datasets, while the latter gives more improvement on CTB6 and PKU datasets. Based on the results of lattice LSTM on the four examined datasets, ``Lattice+Subword'' has a comparable performance with ``Lattice+Word''.

\begin{table}[!tp]
\begin{center}
\resizebox{\columnwidth}{!}{%
\begin{tabular}{|l|l|l|l|l|l|l|}
\hline \textbf{Lexicon vs. Pretrained Emb} &  \textbf{P} & \textbf{R} & \textbf{F1} & \textbf{ER}\% & $\textbf{R}_{IV}$ & $\textbf{R}_{OOV}$\\ \hline
Baseline  &95.93 &95.62 &95.78&0 &96.70 &77.36\\
\hline
Lattice+Subword Rand Emb &96.13  &95.82 &95.97&-4.5 &96.85 &78.37\\
Lattice+Subword Emb &96.23 &95.90  &96.07&-6.9 &96.86 &79.79\\
Lattice+Word Rand Emb &96.26  &96.12 &96.19&-9.7 &97.00 &81.28\\
Lattice+Word Emb &96.36  &96.16 &96.27& -11.6 &97.05 &81.02\\
\hline
\end{tabular}
}
\end{center}
\caption{Lexicon and embeddings on CTB6.}
\label{tab:randemb}
\end{table}

\begin{figure}[!tp] 
  \centering 
  \includegraphics[width=3.2in]{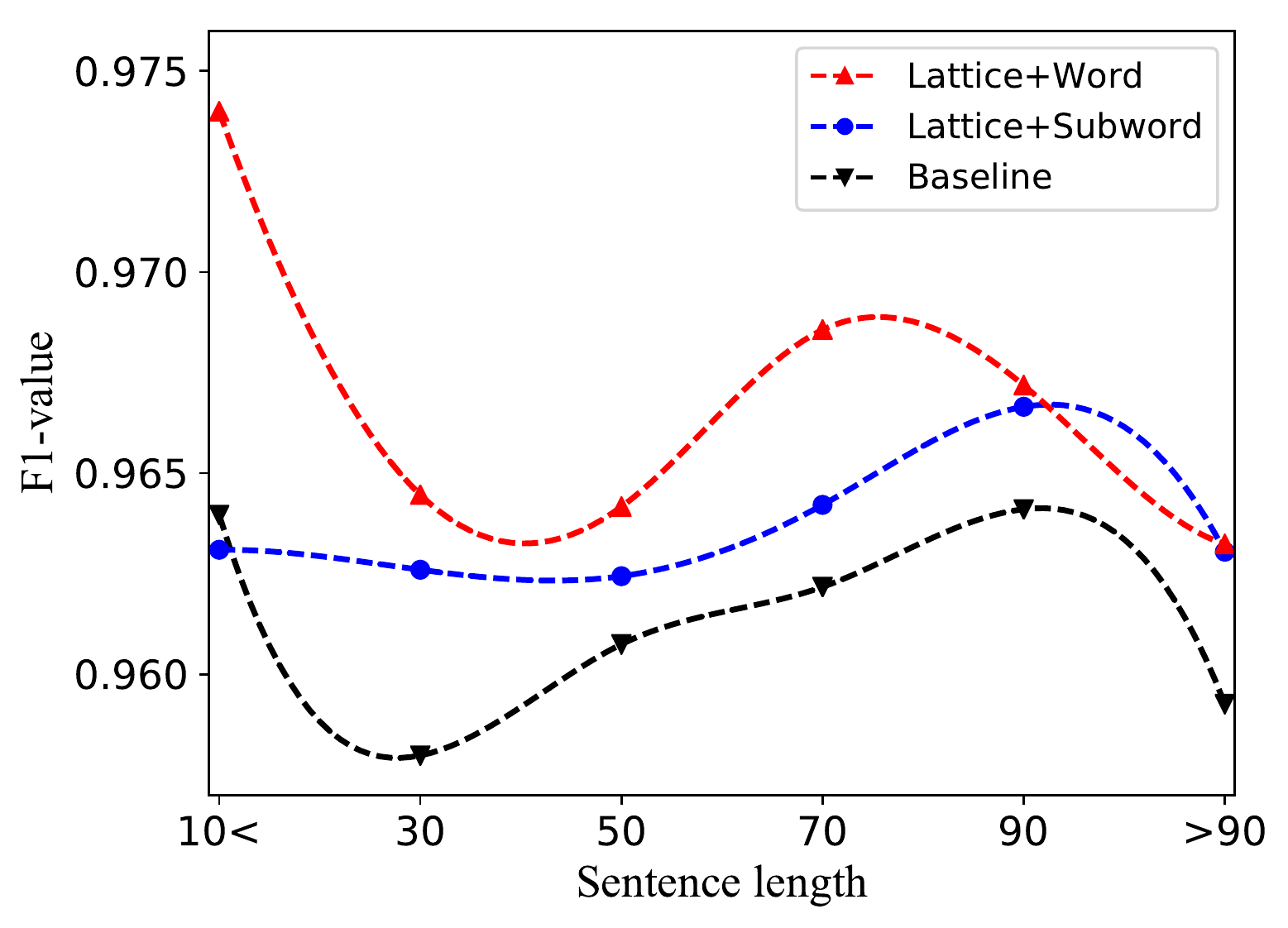}
  \caption{F1-value against the sentence length.}
  \label{fig:sentencelength}
\end{figure}

\subsection{Analysis}
\noindent \textbf{Lexicon and Embeddings}. Table \ref{tab:randemb} shows the model results on CTB6 test data. To distinguish the contribution from word/subword lexicon and their pretrained embeddings, we add another set of experiments by using the same word/subword lexicon with randomly initialized embeddings\footnote{Within $[-\sqrt{\frac{3}{dim}},\sqrt{\frac{3}{dim}}]$, $dim$ is the embedding size.}. As shown in Table \ref{tab:randemb}, the lattice LSTM models consistently outperform the baseline. Lattice LSTM with pretrained word embeddings gives the best result, with an $11.6\%$ error reduction. The word auxiliary lattice LSTM outperforms the model with subword information on CTB6 dataset. The contribution of error reduction by the lexicon in "Lattice+Word" and "Lattice+Subword" are $4.5\%$ and $9.7\%$, respectively. We can estimate the contribution of pretrained embeddings for "Lattice+Word" and "Lattice+Subword" are $(6.9\%-4.5\%) = 2.4\%$ and $(11.6\% - 9.7\%) = 1.9\%$, respectively. The comparison between pretrained embeddings and randomly initialized embeddings shows that both pretrained embeddings and lexicon are useful to lattice LSTM, and lexicon contributes more than the pretrained embeddings. 

\noindent \textbf{OOV Analysis}. Table \ref{tab:randemb} also shows the recall of in-vocabulary ($\textbf{R}_{IV}$) and out-of-vocabulary ($\textbf{R}_{OOV}$) words, respectively. As shown in the table, the recall of out-of-vocabulary words can be largely improved with the lattice structure ($2.43\%, 3.66\%$ absolute improvement for subword encoding and word encoding, respectively). The $\textbf{R}_{OOV}$ of "Lattice+Subword" are largely improved ($+1.42\%$) with the pretrained subword embeddings. It is interesting that the $\textbf{R}_{OOV}$ of "Lattice+Word" has a slight reduction when using pretrained word embeddings, we leave the investigation of this phenomenon in future work.

\noindent \textbf{Sentence Length}. We compare the baseline model, lattice model with subword embeddings and word embeddings based on the sentence length. Figure \ref{fig:sentencelength} shows the F1 distribution on CTB6 dev dataset with respect to sentence length on three models. The baseline The performance of baseline has a trough in around 30-character sentences and decreases when the sentence length over 90, this phenomenon has also been observed in transition-based neural segmentor \citet{yang2017neural}. ``Lattice+Word'' has a similar performance-length curve while "Lattice+Subword" gives a more stable performance along sentence length. One possible reason is that words are built using the auto-segmented corpus whose segmentor has the similar performance-length distribution. On the other hand, subwords in BPE algorithm are built on corpus level statistic information which is not related with sentence length. Hence the "Lattice+Subword" model gives a more stable performance distribution along sentence length.

\begin{table}[!tp]
\begin{center}
\resizebox{\columnwidth}{!}{%
\begin{tabular}{|l|l|l|l|l|l|l|}
\hline 
\textbf{Data}&\textbf{Emb} &\textbf{Split}&\textbf{\#Word}  & \textbf{\#Match} & \textbf{Ratio} (\%) & \textbf{ER} (\%)\\ 
\hline
\multirow{4}*{CTB6} 
&\multirow{2}*{Word}&Train &641k  &573k  &89.35 &--\\
&&Test &81.6k &73.3k &89.79 & 11.9\\
\cline{2-7}
&\multirow{2}*{Subword}&Train &641k   &536k &83.57 &--\\
&&Test & 81.6k   &68.6k &84.13 & 7.14 \\
\hline
\multirow{4}*{PKU} 
&\multirow{2}*{Word}&Train &1.01m  &967k  &95.89 &--\\
&&Test &104k &101k &96.63 & 12.8\\
\cline{2-7}
&\multirow{2}*{Subword}&Train &1.01m   &918k &90.87 &--\\
&&Test & 104k   &95.4k &91.42 &10.6 \\
\hline
\multirow{4}*{MSR} 
&\multirow{2}*{Word}&Train &2.12m  &1.98m  &93.37 &--\\
&&Test &107k &99.8k &93.38 &11.5 \\
\cline{2-7}
&\multirow{2}*{Subword}&Train &2.12m   &1.93m &91.12 &--\\
&&Test & 107k   &98.2k &91.91 & 15.4\\
\hline
\multirow{4}*{Weibo} 
&\multirow{2}*{Word}&Train &421k  &370k  &87.97 &--\\
&&Test &188k &162k &86.03 & 2.0\\
\cline{2-7}
&\multirow{2}*{Subword}&Train &421k   &337k &80.10 &--\\
&&Test & 188k   &147k &78.39 &6.0 \\
\hline
\end{tabular}
}
\end{center}
\caption{Word/Subword coverage in lexicon. \textbf{\#Word} is the number of words in the corresponding dataset, \textbf{\#Match} is the number of matched words between the dataset and word/subword lexicon, $\textbf{\#Ratio}=\frac{\textbf{\#Match}}{\textbf{\#Word}}$ represents the word coverage rate. \textbf{\#ER} is the error reduction compared with baseline model. }
\label{tab:lexiconmatch}
\end{table}

\noindent \textbf{Word/Subword Coverage in lexicon}. Table \ref{tab:lexiconmatch} shows the word/subword coverage rate between word/subword lexicon with four datasets. Word/subword level coverage is consistently higher than the entity level coverage in \citet{zhang18chinese}. In ``Lattice+Word'' and ``Lattice+Subword'' models, higher word/subword coverage (PKU/MSR, $>90\%$) gives better error reduction rate. And on Weibo dataset, both two lattice models have limited improvements, as the word/subword coverage in this data is the lowest. On the other hand, although ``Lattice+Subword'' has lower word coverages in all datasets, it gives better performance than ``Lattice+Word'' on MSR and Weibo datasets. This shows that both the word coverage and the quality of subsequence embeddings are critical to lattice LSTM model and lattice LSTM with subword encoding can give comparable performance with lower word coverage.

\noindent \textbf{Case Study}. Figure \ref{fig:case} shows two examples of the segmentation results on CTB6 test dataset. In example 1, both baseline and ``Lattice+Subword'' fail to give correct segmentation of ``狮子会\;(Lions Clubs)'' while ``Lattice+Word'' can successfully distinguish it. In this example, both ``Lattice+Word'' and ``Lattice+Subword'' have the noisy matched word/subword ``狮子(Lion)'', but there is an extra matched word ``狮子会\;(Lions Clubs)'' in ``Lattice+Word'' to provide a stronger evidence of segmentation due to the gate control mechanism. Example 2 shows another example that ``Lattice+Subword'' gives the right segmentation while ``Lattice+Word'' fails. In this case, both ``Lattice+Word'' and ``Lattice+Subword'' have the correct matched word/subword ``多样性(Diversity)'' while ``Lattice+Word'' has an extra noisy matched word ``性日($\times$)'' which misleads ``Lattice+Word'' to segment the sentence incorrectly. We conclude that the gate control mechanism for matched words/subwords is useful but not perfect. Based on the final performance in table \ref{tab:mainresults}, the lattice LSTM with gate control structure has advantages outweigh its disadvantages.

\begin{figure}[!t] 
 \begin{subfigure}{}\centering
 \resizebox{\columnwidth}{!}{%
  \begin{tabular}{|ll|l|}
  \hline
  \multicolumn{3}{|c|}{\multirow{2}*{\textbf{\#Example 1: where Lattice+Subword fails.} }}\\
  \multicolumn{3}{|c|}{}\\
  \hline
  \multicolumn{2}{|c|}{\multirow{2}*{\textbf{Sentence}}}& 国际狮子会帮助湖北灾民住进新居\\
   & & Int'l Lions Clubs help Hubei flood victims move in new house\\
  \hline
  \multicolumn{2}{|c|}{\multirow{2}*{\textbf{Gold Segmentation}}} &国际 / 狮子会 / 帮助 / 湖北 / 灾民 / 住进 / 新居\\
  &&Int'l/Lions Clubs/help/Hubei/flood victims/move in/new house\\
  \hline
  \hline
  \multicolumn{2}{|c|}{\multirow{2}*{\textbf{Baseline}}} &\colorbox{green!30}{国际/}\colorbox{red!30}{狮子 / 会}\colorbox{green!30}{ / 帮助 / 湖北 / 灾民 / 住进 / 新居}\\
  &&\colorbox{green!30}{Int'l/}\colorbox{red!30}{Lion/will}\colorbox{green!30}{/help/Hubei/victims/move in/new house}\\
  \hline
  \hline
  \multicolumn{1}{|l|}{\multirow{4}*{\textbf{L+Word}}}&\multirow{2}*{\textbf{Matched}} &国际,狮子,\underline{狮子会},帮助,湖北,灾民,灾民住,民住,住进,新居\\
  \multicolumn{1}{|l|}{}&  &Int'l,Lions,\underline{Lions Clubs},help,Hubei,victims,$\times$,$\times$,move in,new house\\
  \cline{2-3}
  \multicolumn{1}{|l|}{}& \multirow{2}*{\textbf{Decode}} &\colorbox{green!30}{国际 / 狮子会 / 帮助 / 湖北 / 灾民 / 住进 / 新居}\\
  \multicolumn{1}{|l|}{}&  &\colorbox{green!30}{Int'l/Lions Clubs/help/Hubei/flood victims/move in/new house}\\
  \hline
  \hline
  \multicolumn{1}{|l|}{\multirow{4}*{\textbf{L+Subword}}}&\multirow{2}*{\textbf{Matched}} &国际,狮子,帮助,湖北,灾民,新居\\
  \multicolumn{1}{|l|}{}&  &Int'l,Lions,help,Hubei,victims,new house\\
  \cline{2-3}
  \multicolumn{1}{|l|}{}& \multirow{2}*{\textbf{Decode}} &\colorbox{green!30}{国际 / }\colorbox{red!30}{狮子 / 会}\colorbox{green!30}{ / 帮助 / 湖北 / 灾民 / 住进 / 新居}\\
  \multicolumn{1}{|l|}{}&  &\colorbox{green!30}{Int'l/}\colorbox{red!30}{Lion/will}\colorbox{green!30}{/help/Hubei/victims/move in/new house}\\
  \hline
  \end{tabular}
  }
  \end{subfigure}
  \begin{subfigure}{}\centering
  \resizebox{\columnwidth}{!}{%
  \begin{tabular}{|ll|l|}
  \hline
  \multicolumn{3}{|c|}{\multirow{2}*{\textbf{\#Example 2: where Lattice+Word fails.} }}\\
  \multicolumn{3}{|c|}{}\\
  \hline
  \multicolumn{2}{|c|}{\multirow{2}*{\textbf{Sentence}}}& 国际生物多样性日纪念大会在京举行\\
   & & Int'l Biological Diversity Day COMM meeting in Beijing hold\\
  \hline
  \multicolumn{2}{|c|}{\multirow{2}*{\textbf{Gold Segmentation}}} &国际 / 生物 / 多样性 / 日 / 纪念 / 大会 / 在 / 京 / 举行\\
  &&Int'l/Biological/Diversity/Day/COMM/meeting/in/Beijing/hold\\
  \hline
  \hline
  \multicolumn{2}{|c|}{\multirow{2}*{\textbf{Baseline}}} &\colorbox{green!30}{国际 / 生物 / }\colorbox{red!30}{多样性日 / }\colorbox{green!30}{纪念 / 大会 / 在 / 京 / 举行}\\
  &&\colorbox{green!30}{Int'l/Biological/}\colorbox{red!30}{DiversityDay}\colorbox{green!30}{COMM/meeting/in/Beijing/hold}\\
  \hline
  \hline
  \multicolumn{1}{|l|}{\multirow{4}*{\textbf{L+Word}}}&\multirow{2}*{\textbf{Matched}} &国际,生物,\underline{多样性},\underline{性日},纪念,大会,在京,京举,举行\\
  \multicolumn{1}{|l|}{}&  &Int'l,Biological,Diversity,\underline{$\times$},COMM,meeting,in Beijing,$\times$,hold\\
  \cline{2-3}
  \multicolumn{1}{|l|}{}& \multirow{2}*{\textbf{Decode}} &\colorbox{green!30}{国际 / 生物 / }\colorbox{red!30}{多样 / 性日}\colorbox{green!30}{ / 纪念 / 大会 / 在 / 京 / 举行}\\
  \multicolumn{1}{|l|}{}&  &\colorbox{green!30}{Int'l/Biological/}\colorbox{red!30}{Diverse/$\times$}\colorbox{green!30}{/COMM/meeting/in/Beijing/hold}\\
  \hline
  \hline
  \multicolumn{1}{|l|}{\multirow{4}*{\textbf{L+Subword}}}&\multirow{2}*{\textbf{Matched}} &国际,生物多样性,\underline{多样性},纪念,大会,在京,举行\\
  \multicolumn{1}{|l|}{}&  &Int'l,Biological Diversity,\underline{Diversity},COMM,meeting,in Beijing,hold\\
  \cline{2-3}
  \multicolumn{1}{|l|}{}& \multirow{2}*{\textbf{Decode}} &\colorbox{green!30}{国际 / 生物 / 多样性 / 日 / 纪念 / 大会 / 在 / 京 / 举行}\\
  \multicolumn{1}{|l|}{}&  &\colorbox{green!30}{Int'l/Biological/Diversity/Day/COMM/meeting/in/Beijing/hold}\\
  \hline
  \end{tabular}
  }
  \end{subfigure}
  \caption{Examples. Red and green color represent incorrect and correct segmentation, respectively. $\times$ represents ungrammatical word. Words with underlines are critical to the segmentation errors.}
  \label{fig:case}
\end{figure}

\section{Conclusion}
We investigated the use of ambiguous words and subwords for CWS with lattice LSTM. Subsequences using word embeddings collected from auto-segmented text and subword embeddings deduced from BPE algorithm are empirically compared. Results on four benchmarks show that the subword encoding works comparable with word encoding in lattice LSTM, both significantly improve the segmentation and give comparable performance to the best systems on all evaluated datasets. Experiments also show that the matched subsequence lexicon contributes more than the pretrained embeddings, this shows the potential of incorporating lattice LSTM structure with domain lexicon for cross-domain sequence labeling. We also observe that higher word/subword coverage leads to larger improvement in both ``Lattice+Word'' and ``Lattice+Subword'' models. With the case study, we find that lattice LSTM with gate control structure is useful but can still fail in some cases, which needs deeper investigation in the future. 

\bibliography{aaai2019}
\bibliographystyle{aaai} 

\end{CJK*}

\end{document}